\documentclass{article}
\usepackage{ijcai25}

\usepackage{times}
\usepackage{soul}
\usepackage{url}
\usepackage[hidelinks]{hyperref}
\usepackage[utf8]{inputenc}
\usepackage[small]{caption}
\usepackage{graphicx}
\usepackage{amsmath}
\usepackage{amsthm}
\usepackage{booktabs}
\usepackage{algorithm}
\usepackage{algorithmic}
\usepackage[switch]{lineno}

\usepackage{multirow}
\usepackage{adjustbox} 
\usepackage{subcaption} 
\usepackage{amssymb}
\usepackage{gensymb} 

\usepackage{xspace}
\makeatletter
\DeclareRobustCommand\onedot{\futurelet\@let@token\@onedot}
\def\@onedot{\ifx\@let@token.\else.\null\fi\xspace}
\def\eg{\emph{e.g}\onedot} 
\def\ie{\emph{i.e}\onedot} 
 
\def\etc{\emph{etc}\onedot}

\makeatother

\title{MVAR: MultiVariate AutoRegressive Air Pollutants Forecasting Model}

\author{
Xu Fan$^{1,*}$
\and
Zhihao Wang$^{2,1,*}$\and
Yuetan Lin$^{1}$\and
Yan Zhang$^{3}$\and
Yang Xiang$^{2,\dagger}$\And
Hao Li$^{3,1,\dagger}$\\
\affiliations
$^1$Shanghai Academy of Artificial Intelligence for Science$^\ddagger$\\
$^2$Tongji University$^\ddagger$\\
$^3$Fudan University\\
\emails
fanxu@sais.com.cn,
zhihaolancorrect@tongji.edu.cn,
linyuetan@sais.com.cn,
yan\_zhang@fudan.edu.cn,
shxiangyang@tongji.edu.cn,
lihao\_lh@fudan.edu.cn
}

\DeclareMathOperator{\sa}{SelfAttention}
\DeclareMathOperator{\soft}{Softmax}
\DeclareMathOperator{\ffd}{FeedForward}
\DeclareMathOperator{\ds}{Downsampler}
\DeclareMathOperator{\layernorm}{LayerNorm}
\DeclareMathOperator{\mean}{mean}
\DeclareMathOperator{\std}{std}

\begin{document}

\maketitle

\renewcommand{\thefootnote}{\fnsymbol{footnote}}
\footnotetext[1]{Contribution Equally}
\footnotetext[2]{Corresponding Author}
\footnotetext[3]{Equal contribution as first institutions.}

\renewcommand{\thefootnote}{\arabic{footnote}} 
\setcounter{footnote}{0} 

\begin{abstract}
Air pollutants pose a significant threat to the environment and human health, thus forecasting accurate pollutant concentrations is essential for pollution warnings and policy-making. Existing studies predominantly focus on single-pollutant forecasting, neglecting the interactions among different pollutants and their diverse spatial responses. To address the practical needs of forecasting multivariate air pollutants, we propose MultiVariate AutoRegressive air pollutants forecasting model (MVAR), which reduces the dependency on long-time-window inputs and boosts the data utilization efficiency. We also design the Multivariate Autoregressive Training Paradigm, enabling MVAR to achieve 120-hour long-term sequential forecasting. Additionally, MVAR develops Meteorological Coupled Spatial Transformer block, enabling the flexible coupling of AI-based meteorological forecasts while learning the interactions among pollutants and their diverse spatial responses.
As for the lack of standardized datasets in air pollutants forecasting, we construct a comprehensive dataset covering 6 major pollutants across 75 cities in North China from 2018 to 2023, including ERA5 reanalysis data and FuXi-2.0 forecast data.
Experimental results demonstrate that the proposed model outperforms state-of-the-art methods and validate the effectiveness of the proposed architecture. Both datasets and codes\footnote{Code and datasets will be released publicly when the paper is accepted.} will be publicly available for advanced researches in air pollutants forecasting.
\end{abstract}

\section{Introduction}
Air pollution consists of physical, biological, and chemical substances released from both anthropogenic and natural sources, posing risks to the environment and human health~\cite{shaddick2020half,meo2024effect,wang2024air}. According to the World Health Organization (WHO), major atmospheric pollutants include particulate matter with diameter less than 10 micrometers (PM\textsubscript{10}) and 2.5 micrometers (PM\textsubscript{2.5}), sulfur dioxide (SO\textsubscript{2}), carbon monoxide (CO), ozone (O\textsubscript{3}), and nitrogen dioxide (NO\textsubscript{2}).
Accurate pollutants forecasting is essential for issuing timely warnings and guiding policy decisions to implement effective control measures.

Air pollutants forecasting methods can be broadly categorized into physics-based methods and data-driven methods. Physics-based methods simulate atmospheric chemical mechanisms and transportation processes by coupling meteorological fields~\cite{kryza2024quantifying,maison2024significant}.
These models have slow computation speeds and face challenges in assimilating pollutants observation data, which limits their forecasting efficiency. In contrast, data-driven methods rely on observed pollutant concentrations, offering lower computational demands, faster execution, and higher accuracy, which is suitable for practical use.

Numerous studies have investigated air pollutants forecasting based on monitoring station data, ranging from urban-scale~\cite{ma2024forecasting} to nationwide forecasts~\cite{Airformer2023}. These studies primarily focus on pollutants with significant spatial transportation characteristics, such as PM\textsubscript{2.5} and PM\textsubscript{10}.
Two dominant approaches have emerged: graph-based networks and attention-based models. Graph-based networks rely on prior knowledge to construct relationships among stations and develop a graph structure. However, spatial transportation relationships vary across pollutants and can change under different meteorological conditions. Consequently, graph-based networks~\cite{zhang2023air,GAGNN2023} are typically limited to single-pollutant forecasting and exhibit poor performance when significant meteorological events occur. In contrast, attention-based models~\cite{zhang2020multi,xie2023multi} demonstrate superior capability in capturing inter-station interactions. These models have shown improved performance compared to graph-based networks. 
But they focus on single-pollutant forecasting, failing to fully exploit their advantages, such as implicitly learning the spatially dynamic responses of different pollutants.
For instance, 
PM\textsubscript{2.5} and PM\textsubscript{10} require consideration of transportation effects~\cite{li2020identification}, and O\textsubscript{3} involves photochemical reactions~\cite{wang2023quantitative}.

In recent years, deep learning based meteorological models, such as PanGu~\cite{pangu2023}, FuXi~\cite{chen2023fuxi,zhong2024fuxi}, and GraphCast~\cite{lam2023learning} have significantly outperformed traditional Numerical Weather Prediction (NWP) methods. These models require fewer computational resources, operate more efficiently, and provide gridded forecasts with higher resolution, offering useful information on regional pollutant transport. The advancements in meteorological models have created new possibilities for AI-based meteorologically coupled air pollutants forecasting.

In this study, we propose MultiVariate AutoRegressive air pollutants forecasting model (MVAR) capable of predicting multiple atmospheric pollutants. This study investigates the implicit learning capabilities in capturing diverse transportation modes for various pollutants. Tailored to real-world forecasting needs, we introduce a novel Multivariate Autoregressive Training Paradigm that narrows the temporal input window while enabling accurate long-term prediction capabilities. Furthermore, we explore the coupling of urban pollutants forecasting with AI-based meteorological prediction, which significantly improves the accuracy of multi-pollutant forecasts. Additionally, by incorporating the greedy algorithm for joint modeling, our approach achieves 120-hour forecasts at a 1-hour resolution, greatly enhancing its practicality and real-world applicability.

Data-driven methods face challenges with missing observational data in practical applications, and the field lacks a benchmark dataset. This study focuses on North China, a region characterized by high industrialization and severe pollution~\cite{meng2024long}, making it highly valuable for research. We design a comprehensive data quality control scheme to generate a high-quality urban-level pollutants dataset, integrating ERA5 data and FuXi-2.0 forecast data. This dataset will be publicly released as a benchmark to advance research and applications in this field.

The main contributions of this study are as follows:
\begin{itemize}
\item We introduce a novel training paradigm that narrows the input window while achieving 120-hour forecasts, enhancing the model's long-term forecasting capability.

\item We explore the coupling of AI-based meteorological models with sparse city-level pollutants forecasting, and achieve the goal of simultaneously improving the forecasting accuracy of various pollutants.

\item We release a new benchmark dataset of air pollutants for North China, along with the ERA5 meteorological dataset and the FuXi-2.0 forecast dataset, laying the foundation for future research.
\end{itemize}

\section{Related Work}


\textbf{Physical model-based methods.}
Physical models simulate the chemical mechanisms and transport processes of air pollutants by coupling with meteorological fields~\cite{daly2007air}. WRF-Chem~\cite{kryza2024quantifying}, CMAQ~\cite{pan2025integrated}, and CHIMERE~\cite{maison2024significant} are three widely used regional air quality and atmospheric chemistry-physics models. These models simulate meteorological dynamics and atmospheric chemical processes, supporting multi-scale air pollution forecasting through the coupling of meteorological models. However, physical models heavily rely on high-quality source emission inventories and are sensitive to the selection of empirical parameterization schemes. Inaccurate initial and boundary conditions can lead to forecast failures. Additionally, physical models are slow in computation and unable to assimilate real-time observational data to improve forecasting performance.

\textbf{Data-driven methods.}
In contrast, data-driven methods~\cite{zhang2022deep,zhang2023multi,liao2024probing} have emerged as the dominant approach, leveraging deep learning models to capture spatiotemporal dependencies in air quality data. These methods can be divided into graph-based and attention-based models. Graph-based models~\cite{huang2021spatio,zhang2023air,wang2024forecasting} primarily rely on prior knowledge to construct spatial relationship graphs among stations or cities. For instance, GAGNN~\cite{GAGNN2023} constructs city and city group graphs to capture spatial and latent dependencies between cities. By employing a differentiable grouping network to discover latent correlations and a message-passing mechanism to model dependencies, GAGNN enables accurate nationwide city-level air quality forecasting. Attention-based models~\cite{zhang2020multi,xie2023multi,ma2024forecasting} rely on attention mechanisms to capture dependencies among stations. AirFormer~\cite{Airformer2023} is a Transformer-based model for nationwide air quality prediction that combines deterministic spatiotemporal learning and stochastic uncertainty modeling to enhance forecasting accuracy. However, these studies often focus on single-pollutant forecasting and overlook the chemical interactions among different pollutants and their spatial responses.

Moreover, data-driven methods typically include historical meteorological information to learn the impact of weather on pollutants. However, those meteorological data are often sparse, making it difficult to capture complex atmospheric processes.
The recent years have witnessed significant breakthroughs in AI-based meteorological models, \eg, FourCastNet~\cite{pathak2022fourcastnet}, Pangu-Weather~\cite{pangu2023}, and FuXi~\cite{chen2023fuxi,zhong2024fuxi}, which surpass traditional NWP models in multiple metrics. Both FuXi-2.0 and Pangu-Weather can provide global forecasts with hourly resolution, laying the foundation for building coupled pollutant-meteorology models.


\section{Methodology}

\subsection{Problem Definition}
Our goal is multi-step, multi-variable air pollutants forecasting. The air pollutant concentration values of $N$ cities at a given time $t$ are denoted as $X_t=\{X_{1,t}, X_{2,t}, ...,X_{i,t},...,X_{N,t}\}$, where $X_{i,t} \in \mathbb{R}^{D}$ represents the air pollutant concentrations of all types for the $i$-th city at time $t$, and $D$ is the number of air pollutant types (\eg, PM\textsubscript{2.5}, PM\textsubscript{10}, \etc). Given the historical air pollutant concentrations for all cities over the past $T$ time steps $X=\{X_t, X_{t-1},..., X_{t-T+1}\}$, we aim to learn a function $F(\cdot)$ that predicts $D$ variables for the next $\tau$ time steps:
\begin{equation}
\hat{X}_{t+1:t+\tau} = F(X)
\end{equation}
where $\hat{X}_{t+1:t+\tau}$ represents the predicted air pollutant concentrations for the future $\tau$ time steps.




\subsection{The Proposed Model}
\begin{figure*}[t] 
\centering
\includegraphics[width=\textwidth]{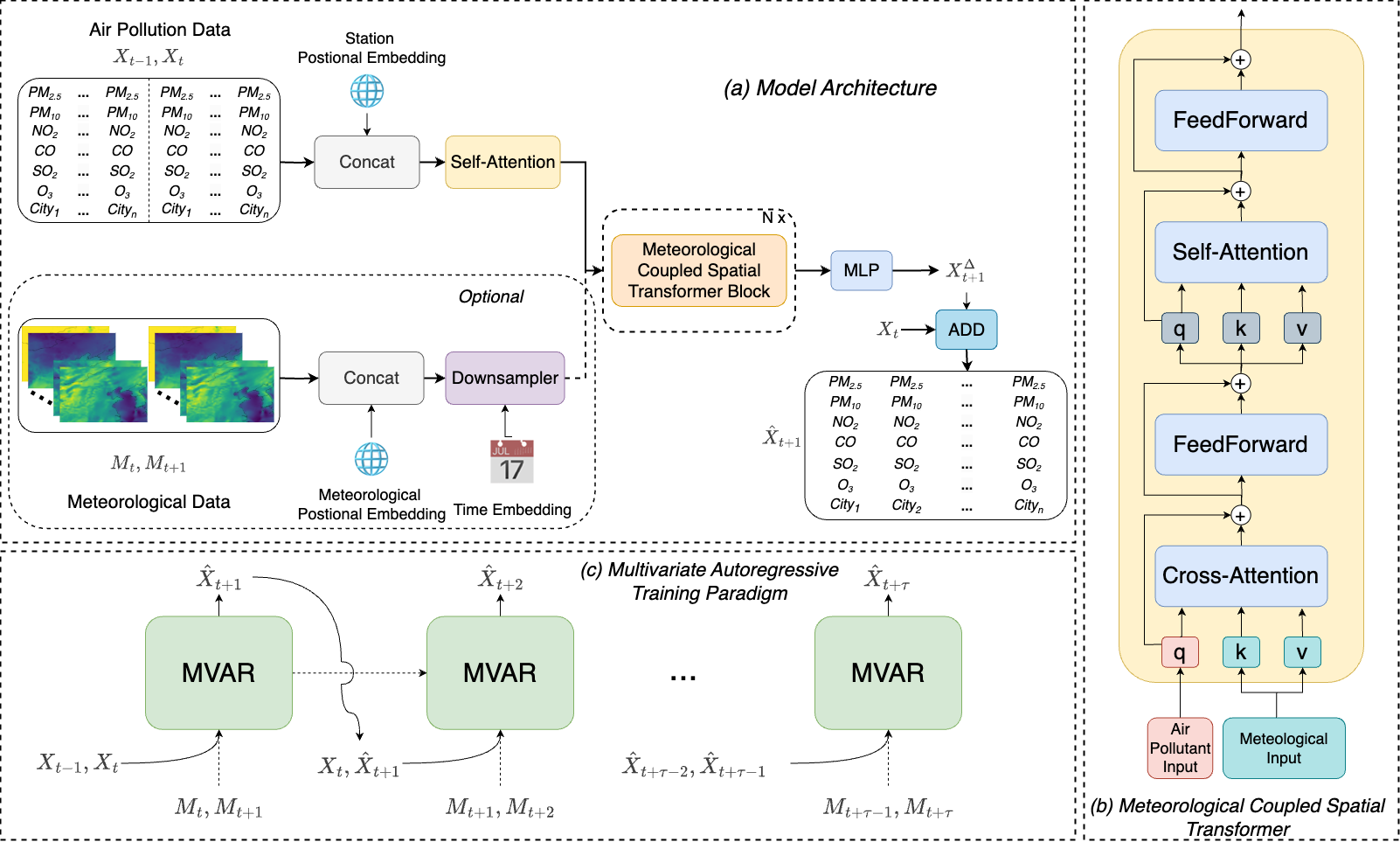} 
\caption{(a) The overall architecture of the proposed model. (b) Meteorological Coupled Spatial Transformer. (c) Multivariate Autoregressive Training Paradigm.} 
\label{model_figure} 
\end{figure*}

Figure \ref{model_figure}(a) illustrates the overall architecture of the model. Given two historical data points \(X_t\) and \(X_{t-1}\), we first map them to high-dimensional hidden vectors and concatenate them with the city position encodings:
\begin{equation}
    \begin{aligned}
        X^{c} &= \sigma(W^a(X_{t-1} \Vert X_{t}) + b^a) \\
        X^{cp} &= X^c \Vert {PE}^a
    \end{aligned}
\end{equation}
where \(\Vert\) denotes concatenation, \(W^a \in \mathbb{R}^{d_{in}\times(2D)}\) and \(b^a \in \mathbb{R}^{d_{in}}\) are trainable parameters, and \(d_{in}\) is the dimension of \(X^c\). \(\sigma\) is the GELU activation function~\cite{hendrycks2016gaussian}, and \({PE}^a \in \mathbb{R}^{N \times d_{pa}}\) denotes the position encodings for the cities, obtained by feeding the latitude and longitude of each city through a fully connected layer. Here, \(d_{pa}\) is the dimension of the position encoding. The concatenated representation \(X^{cp} \in \mathbb{R}^{N \times d_e}\) is formed by combining \(X^c\) and \({PE^a}\), where \(d_e = d_{in} + d_{pa}\). Then, \(X^{cp}\) is input into a self-attention network to learn the interactions among different air pollutants:

\begin{equation}
X^{sa} = \sa(X^{cp}),
\end{equation}
where $\sa$ follows Eq.~\ref{attn1}-\ref{attn4}. However, Eq.~\ref{attn1} is set as follows: $X^q = X^{cp}$, $X^k = X^{cp}$, and $X^v = X^{cp}$. The output ${X}^{sa} \in \mathbb{R}^{N \times d_{in}}$ represents the result of the self-attention mechanism.

For the meteorological data, we use meteorological data at time \(t\) and \(t+1\), denoted as $M_t,M_{t+1} \in \mathbb{R}^{C \times H \times W}$, where \(C\) is the number of meteorological variables, \(H\) is the latitude length, and \(W\) is the longitude length. We concatenate them with the meteorological position encodings to enable spatial alignment in the interaction with air pollutants data:
\begin{equation}
M^{in} = [M_t \Vert M_{t+1} \Vert {PE}^m]
\end{equation}
where ${PE}^m \in \mathbb{R}^{N \times d_{pm}}$ denotes the position encodings for the meteorological data, obtained by feeding the latitude and longitude of each grid point through the same fully connected layer of ${PE}^a$, and $d_{pm}$ is the dimension of the meteorological position encoding. The concatenated representation $M^{in} \in \mathbb{R}^{(2C + d_{pm}) \times H \times W}$ represents the combined meteorological data.

Next, to incorporate temporal information, we add the time encoding to the meteorological data and input the result into a $\ds$ for dimensionality reduction. In this study, the $\ds$ consists of two ResNet blocks, corresponding to two downsampling steps:
\begin{equation}
M^d = \ds(M^{in}, TE)
\end{equation}
where \(TE \in \mathbb{R}^{d_t}\) represents the time encoding, and \(d_t\) denotes the dimension of the time encoding. The downsampled output \(M^d \in \mathbb{R}^{d_e \times H' \times W'}\) is then flattened and reshaped by transposing the last two dimensions, resulting in a final shape of \(N^m \times d_e\), where \(N^m = H' \times W'\).

\subsection{Meteorological Coupled Spatial Transformer}
To account for pollutants transmission characteristics, we design the Meteorological Coupled Spatial Transformer (MCST) block, which enables both the interaction learning of relationships between cities and the coupling of sparse city-level air pollutants data with gridded meteorological forecast data after incorporating meteorological features, as shown in Figure \ref{model_figure}(b). Specifically, the air pollutants data serves as the query, searching for meteorological features affecting its diffusion during the cross-attention stage. In the residual connection, the air pollutants input is added to the output of the attention network to ensure that air pollutants data remains the primary information flow within the model. The detailed process is as follows:

\begin{equation}\label{attn1}
X^q = X^{sa}, \quad X^k = M^{d}, \quad X^v = M^{d}
\end{equation}
\begin{equation}\label{attn2}
Q = X^qW^q, \quad K = X^kW^k, \quad V = X^vW^v
\end{equation}
where $W^q \in \mathbb{R}^{d_{in} \times d_e}, W^k, W^v \in \mathbb{R}^{d_e \times d_e}$ are trainable parameters. Then, we compute the cross-attention scores:
\begin{equation}\label{attn3}
A = \soft\left(\frac{QK^T}{\sqrt{d_e}}\right)
\end{equation}
where $A \in \mathbb{R}^{N \times N^m}$ represents the importance of meteorological features for air pollutants diffusion. Next, the attention result $Z^{ca}  \in \mathbb{R}^{N \times d_e}$ for air pollutants is obtained as:
\begin{equation}\label{attn4}
Z^{ca} = AV
\end{equation}

Finally, the air pollutants input is added to the attention result, and after applying layer normalization, the output is passed via a feedforward module to obtain \(Z^{cf} \in \mathbb{R}^{N \times d_e}\):
\begin{equation}
Z^{cf} = \ffd(\layernorm(Q + Z^{ca}))
\end{equation}
where $\layernorm$ denotes layer normalization. Note that the $\ffd$ module also includes layer normalization, residual connections, and fully connected layers. Similarly, the self-attention performs as follows:
\begin{equation}
    \begin{aligned}
        Z^{sa} =& \sa(Z^{cf}) \\
        Z^{sf} =& \ffd(Z^{cf} + Z^{sa})
    \end{aligned}
\end{equation}
where $\sa$ follows the Eq.~\ref{attn1}-\ref{attn4}, but Eq.~\ref{attn1} is replaced with \(X^q = Z^{cf}, X^k = Z^{cf}, X^v = Z^{cf}\). \(Z^{sa} \in \mathbb{R}^{N \times d_e}\) is the output of the self-attention, and \(Z^{sf} \in \mathbb{R}^{N \times d_e}\) is the output of the $\ffd$ module and also the final output of the MCST. Note that both the cross-attention and self-attention employ the multi-head mechanism.

Assuming that the number of MCST layers is \(L\), we concatenate the outputs of all layers and pass them through two fully connected layers to generate \(X^{\Delta}_{t+1} \in \mathbb{R}^{N \times D}\), which represents the change in concentrations based on \(X_t\). Adding them yields the next-step air pollutant concentrations prediction \(\hat{X}_{t+1} \in \mathbb{R}^{N \times D}\):
\begin{equation}
    \begin{aligned}
        O =& Z^{sf}_1 \Vert Z^{sf}_2 \Vert \cdots \Vert Z^{sf}_j \Vert \cdots \Vert Z^{sf}_L \\
        X^{\Delta}_{t+1} =& W^{o2}(\sigma(W^{o1}O + b^{o1})) \\
        \hat{X}_{t+1} =& X_t + X^{\Delta}_{t+1}
    \end{aligned}
\end{equation}
where \(Z^{sf}_j\) is the output of the \(j\)-th MCST layer, \(O \in \mathbb{R}^{N \times (L \times d_e)}\) is the concatenation of all MCST layer outputs. \(W^{o1} \in \mathbb{R}^{(L \times d_e) \times d_e}\), \(b^{o1} \in \mathbb{R}^{d_e}\), and \(W^{o2} \in \mathbb{R}^{d_e \times D}\) are trainable parameters.

Specifically, if air pollutants data is not coupled with meteorological data, we replace \(X^k\) and \(X^v\) with air pollutants data, \ie, $X^k = X^{sa}, X^v = X^{sa}$, while keeping the remaining steps the same as described above.

\subsection{Multivariate Autoregressive Training Paradigm}
To enable the model to capture the periodic variations of air pollutants, we design the Multivariate Autoregressive Training Paradigm (MATP). As shown in Figure \ref{model_figure}(c), we achieve this by iterating \(\tau\) steps of the proposed model \(f(\cdot)\).

\begin{equation}
F(\cdot) = \underbrace{f(...f(f(\cdot)))}_{\tau \text{ times}}
\end{equation}
When the air pollutants data is not coupled with meteorological features, we present the process \(f(\cdot)\) as
\begin{equation}
\hat{X}_{t+\gamma} = f(\hat{X}_{t+\gamma-2}, \hat{X}_{t+\gamma-1}, \theta), \quad \forall \gamma \in [1, \ldots, \tau]
\end{equation}
where \(\hat{X}_{t} = X_t\), \(\hat{X}_{t-1} = X_{t-1}\), and \(\theta\) denotes the model parameters.
When the air pollutants data is coupled with meteorological features, \(f(\cdot)\) is denoted as:
\begin{equation}
\resizebox{.91\linewidth}{!}{$\hat{X}_{t+\gamma} = f(\hat{X}_{t+\gamma-2}, \hat{X}_{t+\gamma-1}, M_{t+\gamma-1}, M_{t+\gamma}, \theta), \quad \forall \gamma \in [1, \ldots, \tau]$}
\end{equation}

For training, we aim to minimize the error between the predicted values \([\hat{X}_{t+1}, \hat{X}_{t+2}, \ldots, \hat{X}_{t+\tau}]\) and the ground truth values \([X_{t+1}, X_{t+2}, \ldots, X_{t+\tau}]\). In previous studies, the loss functions of models are typically Mean Squared Error (MSE) or Mean Absolute Error (MAE). These loss functions tend to assign larger losses to later steps in the prediction sequence since they are more difficult to predict. This will improve predictions in later steps, but neglect the accuracy of predictions in the early steps. To balance the optimization across all prediction steps, we propose the Step Weighted (SW) loss:

\begin{equation}
\begin{aligned}
\text{SW Loss} &= \frac{1}{\sum_{t=1}^{\tau} w_t} 
\sum_{t=1}^{\tau} w_t \times \\
&\left( \frac{1}{N} \sum_{i=1}^{N} 
\left( \frac{1}{D} \sum_{d=1}^{D} 
\left( \hat{X}_{i, t}^{d} - X_{i, t}^{d} \right)^2 \right) \right)
\end{aligned}
\end{equation}
where \(X_{i,t}^{d}\) represents the concentrations of the $d$-th air pollutant for the \(i\)-th city at time \(t\), and \(w_t\) is the SW loss weight for the \(t\)-th prediction step. The sequence \([w_1, w_2, \ldots, w_{\tau}]\) is a gradually decreasing and equally spaced sequence.

\section{Experiments}

\subsection{Dataset}


The data used in this study include ECMWF ERA5 reanalysis data, FuXi-2.0 model forecast data and ground station measurements from pollution monitoring sites in the North China region\footnote{data available at \url{https://quotsoft.net/air/}}.
The data spans from 2018 to 2023, with 2018–2022 used for training and 2023 as the test set.
Both ERA5 and FuXi datasets are gridded meteorological datasets with a temporal resolution of 1 hour and a spatial resolution of 0.25\degree. We derive FuXi forecast data from FuXi-2.0 model~\cite{zhong2024fuxi}.
In this study, based on the 13-layer output of the FuXi-2.0 model, variables are selected from the 1000 hPa, 925 hPa, and 850 hPa pressure levels, as well as surface.
To address missing values in the pollutant monitoring station data, we propose a data preprocessing scheme. The final dataset includes concentrations of six pollutants across 75 major cities in North China. More detailed data descriptions are provided in the appendix.

\textbf{Data normalization.} In this study, the selected region spans a large area, with significant variations in pollution characteristics and pollutant concentration dynamics among different sub-regions. To enhance the predictive performance of the model, a city-variable approach is used to standardize the concentrations of six pollutants. The standardization process is defined by the following formula:

\begin{align}
X^{norm}_{i,t} = \frac{X_{i,t}-\mean(\tilde{X}_{i})}{\std(\tilde{X}_{i})}
\end{align}
where, $\mean$ and $\std$ represent the mean and standard deviation along the time dimension, respectively. $\tilde{X}_{i}$ denotes the complete historical dataset for the $i$-th city.

\subsection{Implementation Details}
We implement MVAR in PyTorch 2.0.0 using 2 NVIDIA A100 GPUs. We use Adam optimizer. We train 20 epochs with batch size 64 for MVAR, and train 10 epochs with batch size 8 for MVAR with meteorological data. The learning rate is 1e-4 with the weight decay 1e-2. The number of MCST blocks is 3. The hidden dimension in self-attention and MCST is 128 and the number of heads in self-attention and MCST is 4. The maximum and minimum lengths of the SW loss weight sequence are 5 and 0.1. $\tau$ is set to 8.

\subsection{Baselines for Comparison}
We compare our model with the following baselines that fall into the following two categories:
\begin{itemize}
\item Spatial-Temporal Forecasting (STF). Air pollutants forecasting is inherently a spatial-temporal forecasting problem, thus we have selected several highly recognized studies as baselines, \ie, \textbf{STID}~\cite{STID2022}, \textbf{AGCRN}~\cite{AGCRN2020}, \textbf{STGCN}~\cite{STGCN2018} and \textbf{STNorm}~\cite{STNorm2021}. Originally designed for univariate forecasting, these models are adapted to multivariate forecasting in this paper.

\item Air Quality Prediction (AQP). We select 2 strong models for comparison, \ie, \textbf{Airformer}~\cite{Airformer2023} and \textbf{GAGNN}~\cite{GAGNN2023}. Note that these models employ both air pollutants data and meteorological data in their original studies. However, due to the unavailability of station-level and city-level meteorological data, we only employ air pollutants data to train these models in this paper. Similarly, these two models have been adapted to multivariate forecasting.

\end{itemize}

\subsection{Evaluation}
In testing phase, we adopt the same initialization time as in real-world scenarios, specifically at 8:00 and 20:00 (UTC-8) each day, to predict future air pollutant concentrations. We then collect the average performance of these methods for the 1st-4th, 5th-8th, 9th-12th, and 17th-20th prediction steps, respectively. Considering that the time interval for air pollutants data is 6 hours, these steps correspond to the 1-24h, 25-48h, 49-72h, and 97-120h periods, respectively. Root Mean Square Error (RMSE) is used as the evaluation metric. 

In addition, since the original sizes of the baseline models are relatively small, we enlarge their sizes by increasing the hidden dimensions and other parameters to ensure a fair comparison, resulting in model sizes slightly larger than that of the proposed model.

\subsection{Experimental Results}

\begin{table}
    \centering
    \resizebox{\columnwidth}{!}{
    \begin{tabular}{l|cc}
        \hline
        \textbf{Experiment Settings} & \textbf{Train} & \textbf{Test} \\
        \hline
        Past 20 Steps and Next 20 Steps & 5120 & 630 \\
        Past 2 Steps and Next 20 Steps & 5888 & 659 \\
        Past 2 Steps and Next 8 Steps & 6464 & 678 \\
        \hline
    \end{tabular}
    }
    \caption{Statistics for our dataset in three experiment settings. }
    \label{dataset_statistics}
\end{table}

For the SFT and AQP models, predictions for the concentrations of six air pollutants over the next 20 steps are based on the preceding 20 steps, requiring complete data for past 20 steps. In contrast, MVAR predicts the next 20 steps using only the past 2 steps as input, requiring complete data for only the past 2 steps.
Different settings have varying data requirements, resulting in varying amounts of training and testing data, as shown in Table~\ref{dataset_statistics}.
We align the test datasets across all approaches in each experiment.
We design several experiments to validate the effectiveness of MVAR.



\subsubsection{Past 20 Steps and Next 20 Steps}\label{20_20_section}

\begin{table*}[t]
\centering
\setlength{\tabcolsep}{2pt} 
\resizebox{\textwidth}{!}{ 
\begin{tabular}{l|c|c|cccc|cccc|cccc|cccc|cccc|cccc}
\hline 
\multirow{2}{*}{Model} & \multirow{2}{*}{MS (MB)} & \multirow{2}{*}{TDS} & \multicolumn{4}{c|}{SO\textsubscript{2}} & \multicolumn{4}{c|}{NO\textsubscript{2}} & \multicolumn{4}{c|}{PM\textsubscript{2.5}} & \multicolumn{4}{c|}{PM\textsubscript{10}} & \multicolumn{4}{c|}{CO} & \multicolumn{4}{c}{O\textsubscript{3}} \\
\cline{4-7} \cline{8-11} \cline{12-15} \cline{16-19} \cline{20-23} \cline{24-27}
 & & & 1-24h & 25-48h & 49-72h & 97-120h & 1-24h & 25-48h & 49-72h & 97-120h & 1-24h & 25-48h & 49-72h & 97-120h & 1-24h & 25-48h & 49-72h & 97-120h & 1-24h & 25-48h & 49-72h & 97-120h & 1-24h & 25-48h & 49-72h & 97-120h \\
\hline 
STGCN~\cite{STGCN2018} & 3.44 & 5120 & 13.23 & 13.04 & 12.94 & 12.97 & 22.2 & 22.25 & 22.29 & 22.28 & 33.78 & 34.64 & 34.65 & 34.97 & 91.97 & 93.08 & 93.54 & 94.54 & 0.45 & 0.46 & 0.45 & 0.45 & 55.24 & 55.51 & 55.7 & 55.77\\
AGCRN~\cite{AGCRN2020} & 2.95 & 5120 & 17.53 & 16.76 & 17.42 & 16.25 & 25.84 & 25.68 & 25.52 & 25.39 & 42.9 & 42.5 & 43.9 & 41.9 & 104.22 & 103.11 & 102.63 & 103.34 & 0.6 & 0.58 & 0.61 & 0.58 & 52.42 & 52.85 & 52.56 & 52.65 \\
STNorm~\cite{STNorm2021} & 3.10 & 5120 & 13.4 & 13.6 & 13.69 & 13.59 & 21.93 & 22.28 & 22.37 & 22.46 & 34.58 & 35.66 & 35.92 & 36.3 & 91.69 & 93.07 & 93.48 & 94.02 & 0.46 & 0.47 & 0.48 & 0.47 & 52.95 & 52.72 & 52.6 & 52.68\\
STID~\cite{STID2022} & 3.14 & 5120 & 13.01 & 13.12 & 13.1 & 13.08 & 22.36 & 22.38 & 22.4 & 22.49 & 34.93 & 35.31 & 35.44 & 35.82 & 93.24 & 93.82 & 94.13 & 94.61 & 0.46 & 0.46 & 0.46 & 0.46 & 56.07 & 55.93 & 55.85 & 55.72\\
\hline 
Airformer~\cite{Airformer2023} & 3.31 & 5120 & 13.4 & 13.35 & 13.25 & 13.18 & 20.55 & 20.38 & 20.33 & 20.27 & 36.43 & 36.32 & 36.17 & 36.13 & 93.98 & 93.76 & 93.75 & 93.8 & 0.47 & 0.46 & 0.46 & 0.46 & 38.09 & 37.81 & 37.93 & 37.83 \\
GAGNN~\cite{GAGNN2023} & 3.07 & 5120 & 16.34 & 15.64 & 15.64 & 15.55 & 25.07 & 24.77 & 24.56 & 24.49 & 40.34 & 39.58 & 40.63 & 40.32 & 98.12 & 98.62 & 97.36 & 99.92 & 0.58 & 0.56 & 0.57 & 0.57 & 49.74 & 48.98 & 49.72 & 49.94 \\
\hline 
MVAR (ours) & 2.53 & 6464 & \underline{12.34} & \underline{12.3} & \underline{12.35} & \underline{12.3} & \underline{17.13} & \underline{18.4} & \underline{18.68} & \underline{18.78} & \underline{29.66} & \underline{33.88} & \underline{34.3} & \underline{34.1} & \underline{86.19} & \underline{92.29} & \underline{93.29} & \underline{93.38} & \underline{0.4} & \underline{0.42} & \underline{0.43} & \underline{0.43} & \underline{28.76} & \underline{32.39} & \underline{34.03} & \underline{34.95} \\
MVAR\textsubscript{fuxi} (ours) & 5.93 & 6464 & \textbf{12.14} & \textbf{12.09} & \textbf{12.16} & 12.22 & \textbf{16.07} & \textbf{16.79} & \textbf{16.92} & 17.15 & \textbf{28.09} & \textbf{30.66} & \textbf{30.97} & 31.36 & 84.6 & 89.42 & 90.28 & \textbf{90.15} & \textbf{0.39} & \textbf{0.4} & \textbf{0.41} & \textbf{0.41} & 25.98 & 27.97 & 28.28 & 28.86\\
MVAR\textsubscript{era5} (ours) & 5.93 & 6464 & \textbf{12.14} & 12.11 & 12.18 & \textbf{12.21} & 16.07 & 16.81 & 16.92 & \textbf{16.97} & 28.11 & 30.69 & 31.0 & \textbf{31.29} & \textbf{84.55} & \textbf{89.28} & \textbf{90.17} & 90.47 & \textbf{0.39} & 0.41 & \textbf{0.41} & 0.42 & \textbf{25.93} & \textbf{27.81} & \textbf{28.07} & \textbf{28.06}\\
\hline 
\end{tabular}
}
\caption{Performance comparison of different models in the Past 20 Steps and Next 20 Steps setting. The bold and underlined font mean the best overall result and the best result excluding from MVAR\textsubscript{era5} and MVAR\textsubscript{fuxi}, respectively. MS represents model size and its magnitude is MB. TDS indicates the training dataset size.}
\label{20input_20output_model_performance}
\end{table*}

We first compare the performance of our model with baseline models under the Past 20 Steps and Next 20 Steps setting.

As shown in Table~\ref{20input_20output_model_performance}, all models, except for MVAR\textsubscript{era5} and MVAR\textsubscript{fuxi}, use only air pollutants data for training. MVAR\textsubscript{era5} and MVAR\textsubscript{fuxi} represent models trained with both meteorological and air pollutants data, where ERA5 meteorological data is used during the training phase. During the testing phase, MVAR\textsubscript{era5} uses ERA5 meteorological data, while MVAR\textsubscript{fuxi} uses FuXi-2.0 forecasting data which is suitable for real-world scenarios.

From Table~\ref{20input_20output_model_performance}, we observe that MVAR consistently outperforms all baseline models, indicating that using historical data from merely 2 time steps as input suffices to surpass the performance of models that rely on much longer historical sequences for input. This indicates that relying on long historical input is not essential. Instead, choosing shorter historical windows based on periodicity empowers the model to capture periodic patterns effectively. By comparing the STF models with MVAR, we find that while STF demonstrates good generalization performance in air pollutants prediction, MVAR still outperforms them, which highlights the importance of domain knowledge for model building. Moreover, MVAR significantly surpasses the state-of-the-art (SOTA) transformer-based AQP model AirFormer, which demonstrats that our model effectively captures the periodic variations in air pollutant concentrations and holds an edge over direct-output prediction methods.

We further observe that the incorporation of meteorological data significantly bolsters the performance of pollutants forecasting, with the most remarkable improvements in PM\textsubscript{2.5}, PM\textsubscript{10}, O\textsubscript{3}, and NO\textsubscript{2}. Meteorological factors influence the convective and advective transport processes of PM\textsubscript{2.5} and PM\textsubscript{10}, thereby affecting their local concentrations. Similarly, O\textsubscript{3} and NO\textsubscript{2} are key participants in atmospheric photochemical reactions, where the temperature within the boundary layer plays a crucial role in modulating the reaction rates of these processes. Taking PM\textsubscript{2.5} for example, compared with MVAR, MVAR\textsubscript{era5} reduces RMSE by 5.23\%, 9.42\%, 9.62\%, and 8.24\% for the four future periods, respectively.


\subsubsection{Past 2 Steps and Next 20 Steps}\label{2_20_section}

\begin{table*}[t]
\centering
\setlength{\tabcolsep}{2pt} 
\resizebox{\textwidth}{!}{ 
\begin{tabular}{l|c|c|cccc|cccc|cccc|cccc|cccc|cccc}
\hline 
\multirow{2}{*}{Model} & \multirow{2}{*}{MS (MB)} & \multirow{2}{*}{TDS} & \multicolumn{4}{c|}{SO\textsubscript{2}} & \multicolumn{4}{c|}{NO\textsubscript{2}} & \multicolumn{4}{c|}{PM\textsubscript{2.5}} & \multicolumn{4}{c|}{PM\textsubscript{10}} & \multicolumn{4}{c|}{CO} & \multicolumn{4}{c}{O\textsubscript{3}} \\
\cline{4-7} \cline{8-11} \cline{12-15} \cline{16-19} \cline{20-23} \cline{24-27}
 & & & 1-24h & 25-48h & 49-72h & 97-120h & 1-24h & 25-48h & 49-72h & 97-120h & 1-24h & 25-48h & 49-72h & 97-120h & 1-24h & 25-48h & 49-72h & 97-120h & 1-24h & 25-48h & 49-72h & 97-120h & 1-24h & 25-48h & 49-72h & 97-120h \\
\hline 
AGCRN~\cite{AGCRN2020} & 2.95 & 5888 & 16.46 & 15.73 & 16.33 & 15.33 & 25.1 & 25.06 & 24.8 & 24.76 & 41.44 & 41.02 & 42.83 & 40.86 & 104.01 & 99.16 & 98.77 & 99.26 & 0.57 & 0.55 & 0.58 & 0.55 & 50.66 & 51.33 & 51.08 & 50.81 \\
STID~\cite{STID2022} & 3.03 & 5888 & 13.11 & 13.17 & 13.12 & 13.14 & 22.67 & 22.78 & 22.81 & 22.82 & 35.72 & 36.36 & 36.69 & 36.96 & 96.43 & \underline{93.62} & 93.75 & 94.01 & 0.47 & 0.47 & 0.47 & 0.47 & 54.42 & 54.15 & 54.04 & 54.07 \\
GAGNN~\cite{GAGNN2023} & 3.02 & 5888 & 15.53 & 16.07 & 15.57 & 15.78 & 25.14 & 24.61 & 24.64 & 24.52 & 41.5 & 39.9 & 41.16 & 41.79 & 100.89 & 96.31 & 96.16 & 96.37 & 0.57 & 0.56 & 0.59 & 0.59 & 49.68 & 49.05 & 49.32 & 48.94 \\
\hline 
MVAR (ours) & 2.53 & 6464 & \underline{12.33} & \underline{12.24} & \underline{12.31} & \underline{12.4} & \underline{17.2} & \underline{18.53} & \underline{18.86} & \underline{19.05} & \underline{30.24} & \underline{34.44} & \underline{34.92} & \underline{34.96} & \underline{92.54} & 93.97 & \underline{92.98} & \underline{92.5} & \underline{0.4} & \underline{0.43} & \underline{0.43} & \underline{0.44} & \underline{28.48} & \underline{32.08} & \underline{33.7} & \underline{34.69} \\
MVAR\textsubscript{fuxi} (ours) & 5.93 & 6464 & \textbf{12.12} & \textbf{12.03} & \textbf{12.1} & \textbf{12.27} & \textbf{16.13} & \textbf{16.87} & \textbf{17.0} & 17.3 & \textbf{28.62} & \textbf{31.11} & \textbf{31.39} & 31.73 & 90.73 & 90.64 & 89.64 & \textbf{88.99} & \textbf{0.39} & \textbf{0.41} & \textbf{0.41} & \textbf{0.42} & 25.74 & 27.69 & 28.02 & 28.67\\
MVAR\textsubscript{era5} (ours) & 5.93 & 6464 & 12.13 & 12.04 & 12.13 & \textbf{12.27} & 16.15 & 16.91 & 17.01 & \textbf{17.07} & 28.64 & 31.16 & 31.44 & \textbf{31.56} & \textbf{90.68} & \textbf{90.52} & \textbf{89.56} & 89.3 & \textbf{0.39} & \textbf{0.41} & \textbf{0.42} & \textbf{0.42} & \textbf{25.71} & \textbf{27.59} & \textbf{27.84} & \textbf{27.87} \\
\hline 
\end{tabular}
}
\caption{Performance comparison of different models in the Past 2 Steps and Next 20 Steps setting.}
\label{2input_20output_model_performance}
\end{table*}

For fair comparison, baseline models are configured to use the same input as our model, \ie, historical data from only 2 time steps, but predict air pollutants for next 20 steps. Note that STGCN and STNorm utilize dilated convolutions along the temporal dimension, which is incompatible with input sequences with only 2 steps. Therefore, these two models are excluded. AirFormer is also excluded, since it relies on reconstruction loss and requires the same input-output sequence length.

As shown in Table~\ref{2input_20output_model_performance}, MVAR achieves SOTA performance for all pollutants except for PM\textsubscript{10}. Incorporating meteorological data significantly improves the predictions for PM\textsubscript{2.5}, O\textsubscript{3}, and NO\textsubscript{2}, further solidifying our model's overall performance.
The reason MVAR does not achieve SOTA performance for PM\textsubscript{10} is that PM\textsubscript{10} is a spike-driven pollutant, meaning its concentration can increase sharply. MVAR is less responsive to such abrupt changes in pollutant levels.

\subsubsection{Past 2 Steps and Next 8 Steps}

\begin{table*}[t]
\centering
\setlength{\tabcolsep}{2pt} 
\resizebox{\textwidth}{!}{ 
\begin{tabular}{l|c|c|cc|cc|cc|cc|cc|cc}
\hline 
\multirow{2}{*}{Model} & \multirow{2}{*}{MS (MB)} & \multirow{2}{*}{TDS} & \multicolumn{2}{c|}{SO\textsubscript{2}} & \multicolumn{2}{c|}{NO\textsubscript{2}} & \multicolumn{2}{c|}{PM\textsubscript{2.5}} & \multicolumn{2}{c|}{PM\textsubscript{10}} & \multicolumn{2}{c|}{CO} & \multicolumn{2}{c}{O\textsubscript{3}} \\
\cline{4-5} \cline{6-7} \cline{8-9} \cline{10-11} \cline{12-13} \cline{14-15}
 & & & 1-24h & 25-48h & 1-24h & 25-48h & 1-24h & 25-48h & 1-24h & 25-48h & 1-24h & 25-48h & 1-24h & 25-48h \\
\hline 
AGCRN~\cite{AGCRN2020} & 2.93 & 6464 & 15.31 & 14.74 & 24.84 & 24.79 & 40.76 & 41.07 & 102.28 & 96.81 & 0.57 & 0.57 & 50.21 & 50.62 \\
STID~\cite{STID2022} & 2.93 & 6464 & 13.12 & 13.12 & 22.6 & 22.78 & 35.66 & 36.46 & 95.46 & \underline{92.92} & 0.47 & 0.47 & 54.19 & 53.98 \\
GAGNN~\cite{GAGNN2023} & 3.01 & 6464 & 15.51 & 15.48 & 24.43 & 24.54 & 39.89 & 39.99 & 99.14 & 95.47 & 0.55 & 0.55 & 48.93 & 49.09 \\
\hline 
MVAR (ours) & 2.53 & 6464 & \underline{12.34} & \underline{12.23} & \underline{17.2} & \underline{18.57} & \underline{30.18} & \underline{34.55} & \underline{91.44} & 93.1 & \underline{0.4} & \underline{0.43} & \underline{28.32} & \underline{31.86} \\
MVAR\textsubscript{fuxi} (ours) & 5.93 & 6464 & \textbf{12.13} & \textbf{12.01} & \textbf{16.13} & \textbf{16.89} & \textbf{28.54} & \textbf{31.14} & 89.64 & 89.76 & \textbf{0.39} & \textbf{0.41} & 25.57 & 27.51 \\
MVAR\textsubscript{era5} (ours) & 5.93 & 6464 & 12.14 & 12.02 & 16.14 & 16.94 & 28.57 & 31.2 & \textbf{89.6} & \textbf{89.63} & \textbf{0.39} & \textbf{0.41} & \textbf{25.54} & \textbf{27.41} \\
\hline 
\end{tabular}
}
\caption{Performance comparison of different models in the Past 2 Steps and Next 8 Steps setting.}
\label{2input_8output_model_performance}
\end{table*}

To further align the amount of training data, we design the following experiment.
The baseline models are configured to use the same input as our model, \ie, 2-step historical data and 8-step prediction. Note that the baselines do not use the autoregressive training paradigm, so their performance can only be evaluated for 1-24h and 25-48h future predictions.
Similar to the experiment in previous section, STGCN, STNorm, and AirFormer are excluded due to their input and output constraints.

The experimental results are shown in Table~\ref{2input_8output_model_performance}. With consistent training data, MVAR achieves SOTA performance for all pollutants except for PM\textsubscript{10}. With meteorological data, MVAR further improves the forecasting accuracy for PM\textsubscript{2.5}, O\textsubscript{3}, and NO\textsubscript{2}, confirming it as the overall SOTA.

\subsection{Ablation Study}
In this section, we conduct ablation studies to investigate: (a) the differences between univariate and multivariate forecasting, and (b) the impact of different loss strategies. All experiments are performed without meteorological coupling, with a forecasting time resolution of 6 hours.

\subsubsection{Effect of Multivariate Prediction}

\begin{figure}[t] 
\centering
\includegraphics[width=\columnwidth]{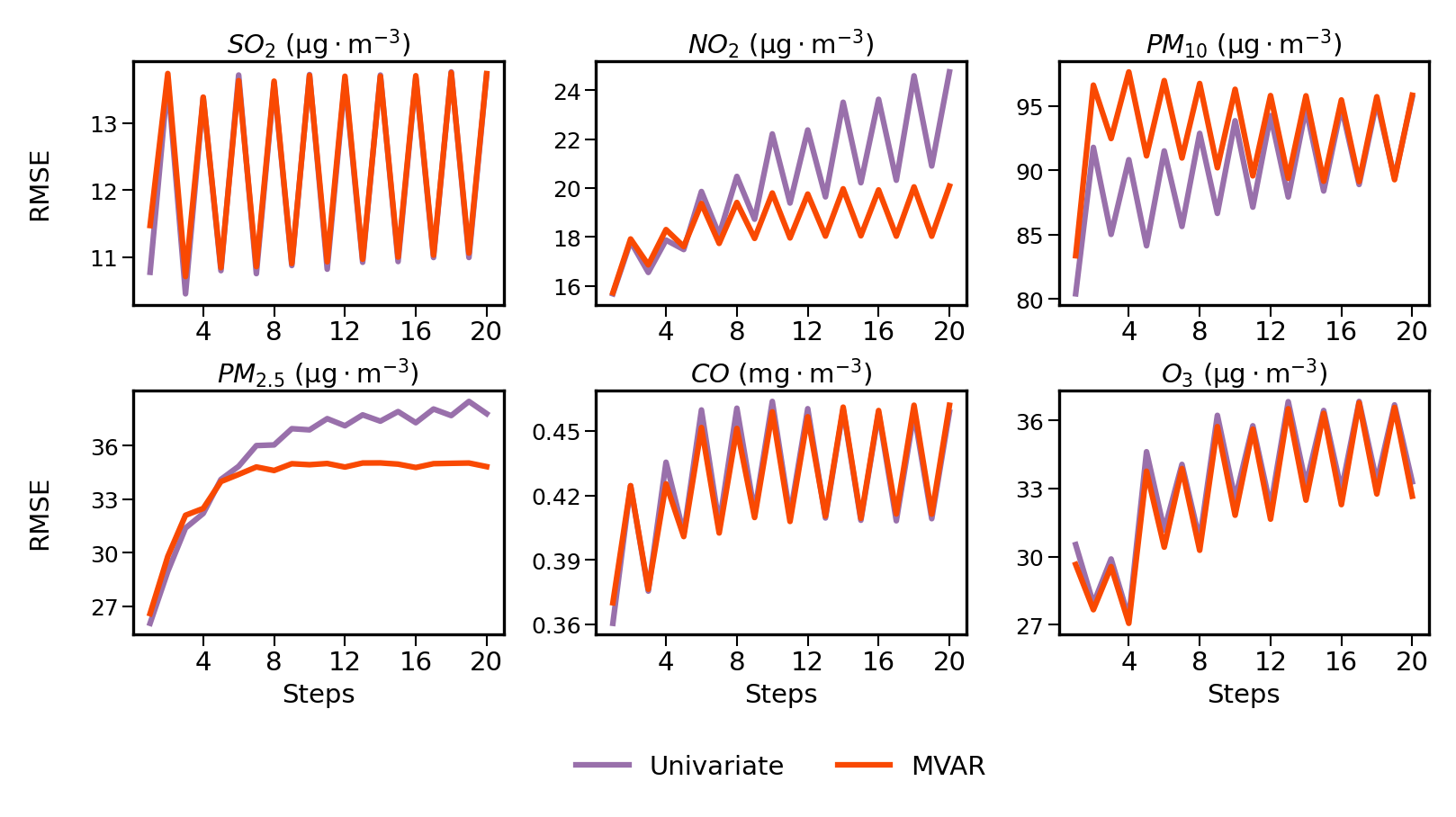} 
\caption{RMSE trend curve of the effect of  Multivariate Prediction.} 
\label{single_vs_multi_rmse_curve} 
\end{figure}

In this study, we employ a multivariate forecasting approach to fully capture the interactions among different air pollutants. To validate the effectiveness of this approach, we train and compare univariate and multivariate forecasting models. The univariate model involves single-pollutant input and output for each pollutant, maintaining consistent model structure and training methods. The comparison results are displayed in Figure~\ref{single_vs_multi_rmse_curve}.
SO\textsubscript{2} and CO are primarily influenced by local emissions and exhibit relatively stable chemical properties in the atmosphere, leading to consistent forecast performance in both univariate and multivariate models. Local emissions are a key factor affecting PM\textsubscript{2.5}, and SO\textsubscript{2} and CO effectively represent the intensity of local emissions, guiding the long-term forecasting of PM\textsubscript{2.5}. Therefore, in multivariate forecasting, PM\textsubscript{2.5} achieves improved prediction accuracy.
Beyond local emissions, photochemical reactions involving O\textsubscript{3} also significantly impact NO\textsubscript{2} concentrations. Long-term univariate forecasts typically show poor performance in later stages for NO\textsubscript{2}. With O\textsubscript{3}, multivariate forecasts convey information about photochemical reactions to the model, enhancing its accuracy. Additionally, changes in SO\textsubscript{2} and CO effectively guide the model's response to local NO\textsubscript{2} emissions, thus improving NO\textsubscript{2} forecast results.

\subsubsection{Effect of Step Weighted Loss Strategies}

\begin{table}
    \centering
    \begin{tabular}{l|ll}
        \hline
        \textbf{Train paradigm} & \textbf{TDS} & \textbf{Epochs} \\
        \hline
        Single-step Pretrain (SP) & 6930 & 20 \\
        SP + 20-step Finetune (SP+20F) & 5120 & 20+10 \\
        20-step Train (20T) & 5120 & 20 \\
        8-step Train (8T) & 6464 & 20 \\
        8-step Weight (8W) & 6464 & 20 \\
        \hline
    \end{tabular}
    \caption{The amount of training data and the number of epochs corresponding to different training paradigms.}
    \label{training_paradigm_statistics}
\end{table}

\begin{figure}[t] 
\centering
\includegraphics[width=\columnwidth]{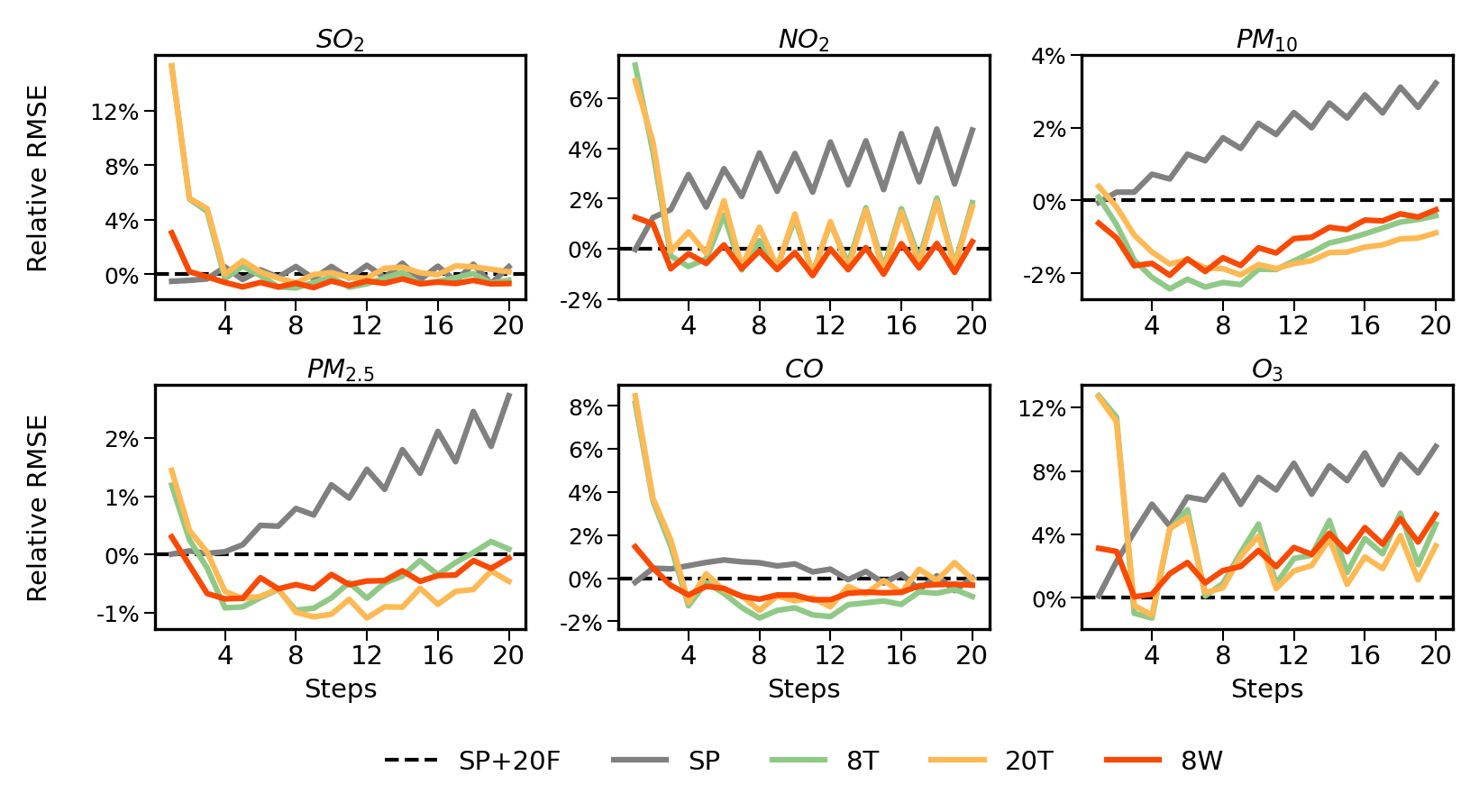} 
\caption{Relative RMSE trend curve of different training paradigms.} 
\label{training_paradigm_relative_rmse_curve} 
\end{figure}

To validate the effectiveness of MATP, we conduct comparative experiments, as shown in Table~\ref{training_paradigm_statistics}. \textbf{SP+20F} represents conducting a 20-step multi-step finetuning based on the single-step model. \textbf{8T} and \textbf{8W} (\ie MATP) refer to training for 8 steps using MAE loss and SW loss, respectively. For ease of comparison, we calculate the relative RMSE of other strategies based on \textbf{SP+20F}, and the results are shown in Figure~\ref{training_paradigm_relative_rmse_curve}. \textbf{8T} and \textbf{20T} exhibit poorer performance in early-stage predictions, particularly for SO\textsubscript{2} and O\textsubscript{3}. However, these strategies improve in later stages as the models focus on optimizing long-term predictions. In contrast, MVAR requires only 8-step training to achieve comparable performance with \textbf{8T} and \textbf{20T} in long-term predictions. At the same time, its early-stage prediction results are very close to those of \textbf{SP+20F}, significantly enhancing data utilization efficiency while reducing training complexity.

\subsection{1-Hour Resolution Air Pollutants Prediction Application}
In the previous experiments, air pollutant concentrations are forecasted at 6-hour intervals, reflecting a coarse temporal resolution. However, real-world applications demand finer resolutions, such as 1-hour intervals, for more precise predictions. To address this, we train four single-step prediction models with lead times of 1h, 3h, 6h, and 24h and utilize a greedy algorithm to achieve 1-hour interval predictions. Results and details are provided in the appendix.

\section{Conclusion}
In this study, MVAR addresses the challenges of multivariate air pollutants forecasting. Using the global hourly-resolution meteorological forecasts generated by FuXi-2.0, for the first time, this work explores coupling sparse urban pollutant data with AI-based meteorological forecast data.
According to the experimental results, the proposed model effectively captures the diverse spatial responses among different pollutants under various meteorological conditions and transmission mechanisms.
And it achieves an ultra-long forecasting time span of 120 hours, setting a new benchmark in the field of air pollutants forecasting.
This model significantly improves the forecasting accuracy of various pollutants. 
A long-term time-series dataset covering 75 cities in North China and 6 major air pollutants have been constructed, integrating ERA5 reanalysis data and FuXi-2.0 forecast data. This standardized dataset provides a solid foundation for fair evaluation and comparative research in the field of air pollutants forecasting.

Possible future work includes delving deeper into the model's  generalization abilities across diverse regions and various meteorological scenarios. Moreover, we will endeavor to boost its application potential by incorporating larger-scale multi-source data.

\bibliographystyle{named}
\bibliography{ijcai25}

\appendix
\section{Appendix}
\subsection{Dataset}
The data used in this study include meteorological data, \ie, ECMWF ERA5 reanalysis data and FuXi-2.0 model forecast data, and air pollutant data, \ie, ground station measurements from pollution monitoring sites in the North China region\footnote{data available at \url{https://quotsoft.net/air/}}.
Air pollution occurs primarily in the Atmospheric Boundary Layer (ABL), where changes in temperature and humidity influence the generation and transformation of pollutants, and wind speed and direction affect the transport of regional pollutants.
Therefore, variables are selected from the 1000 hPa, 925 hPa, and 850 hPa pressure levels, as well as surface-level parameters, based on the 13-layer output of the FuXi-2.0 model.
These variables are shown in Table~\ref{meteorological_variables}.

\begin{table}[ht]
\centering
\begin{tabular}{ccc}  
\toprule  
\textbf{Full Name} & \textbf{Abbreviation} & \textbf{Type} \\  
\midrule  
temperature & T & upper-air \\
u component of wind & U & upper-air \\
v component of wind & V & upper-air \\
specific humidity & Q & upper-air \\
2-meter temperature & T2M & surface \\
2-meter dewpoint temperature & D2M & surface \\
10-meter u wind component & U10 & surface \\
10-meter v wind component & V10 & surface \\
100-meter u wind component & U100 & surface \\
100-meter v wind component & V100 & surface \\
total precipitation & TP & surface \\
\bottomrule  
\end{tabular}
\caption{A summary of all the meteorological variables}
\label{meteorological_variables}
\end{table}

The meteorological data forecasted by FuXi-2.0 model provides information on advective and convective transport conditions through wind speed, while near-surface temperature and dew point temperature offer insights into the photochemical reaction rates near the surface. ERA5 reanalysis data, generated through assimilation of observational data, is considered the most accurate meteorological dataset but does not provide forecasted values. In contrast, FuXi-2.0 offers forecast data that can be directly used for prediction, making it more suitable for real-world applications.

\begin{figure}[t] 
\centering
\includegraphics[width=\columnwidth]{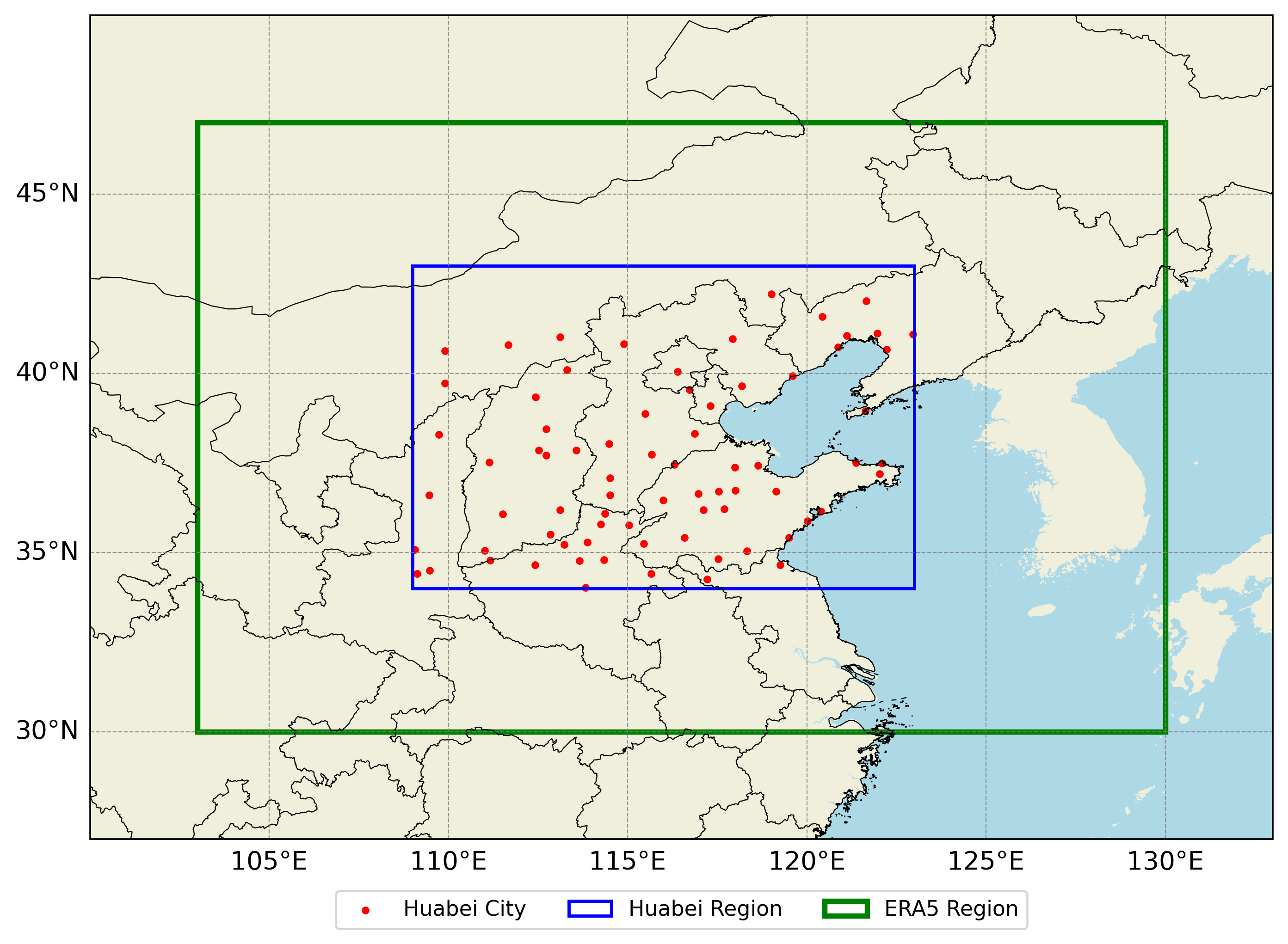} 
\caption{Distribution of 75 North China (Huabei) cities and ERA5 reanalysis data region.} 
\label{distribution_map} 
\end{figure}

Air pollutants data from ground stations in the North China region is also used. However, due to the closure of monitoring stations as a result of urban development, the data contains missing values, anomalies, and gaps, which seriously affects data reliability. To ensure data quality, this study proposes a quality control strategy specifically for multi-city air pollution forecasting. Pollution monitoring stations are concentrated in urban areas, with multiple stations per city, and urban pollution forecasting focuses on the maximum pollutant concentrations within each city. This study applies a three-step quality control process to the ground station data. 

\textbf{The first step} is station-data removal, where stations with more than 50\% missing data are excluded to eliminate stations with poor data quality or those that have stopped operating. \textbf{The second step} is station interpolation, where Kriging interpolation is applied to fill missing data for cities with less than 20\% missing data at a given time.
If the proportion of missing data in a certain time step exceeds 20\%, the data of that time step is excluded.
\textbf{The third step} is city-level aggregation, in which the maximum pollutant concentrations within each city are selected to represent the city's pollution level. This approach addresses the issue of sparse or missing data at individual stations and facilitates more accurate forecasting of the maximum urban pollutant concentrations. This yields a dataset for 75 major cities in North China and the amount of valid data increases from 0 initially to 52,584$\times$75$\times$6 ultimately. The distribution of monitoring stations and the region covered by the meteorological data are shown in Figure~\ref{distribution_map}.


\subsection{Application - 1-Hour Resolution Air Pollutants Prediction}
In the experiments conducted in previous sections, we forecast air pollutant concentrations at 6-hour intervals, representing a coarse-grained temporal resolution. However, in real-world scenarios, we aim for a finer temporal resolution of 1 hour to achieve more precise predictions. Therefore, similar to Pangu-Weather~\cite{pangu2023}, we adopt a greedy algorithm to achieve 1-hour interval predictions for air pollutant concentrations.

Specifically, we train four single-step prediction models with lead times (the time difference between input and output) of 1h, 3h, 6h, and 24h, respectively. We refer to our model in this setting as MVAR\textsubscript{1hour}. Unlike Pangu-Weather, MVAR\textsubscript{1hour} uses the input strategy proposed in this work, \ie, we use only 2-step historical data as inputs. For instance, the input for the single-step prediction model with 1-hour lead time is $X_{t-1}$ and $X_t$, and it outputs $X_{t+1}$. Similarly, for the 3-hour lead time model, the input is $X_{t-3}$ and $X_{t}$, and the output is $X_{t+3}$.

During the prediction process, we iteratively invoke the four single-step prediction models, using the prediction result of the previous step as the input for the next step. A greedy algorithm is employed to reduce accumulated errors. For example, for an 80-hour forecast, the 24-hour prediction model is applied three times, the 6-hour model once, and the 1-hour model 2 times.

\subsubsection{RMSE Trend Curve}
\begin{figure}[t] 
\centering
\includegraphics[width=\columnwidth]{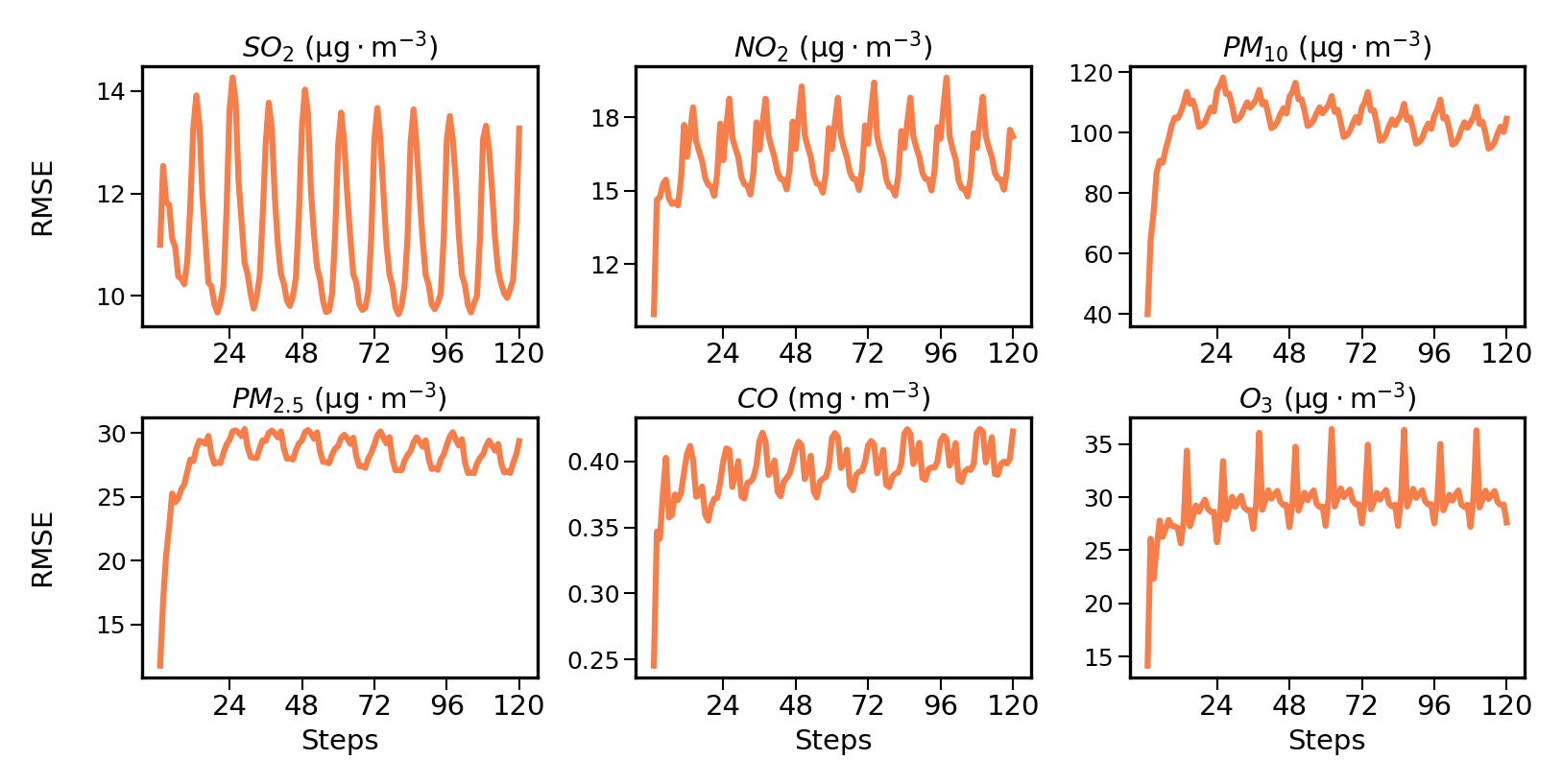} 
\caption{RMSE Trend Curve of MVAR\textsubscript{1hour}} 
\label{pangu_rmse_curve} 
\end{figure}

Figure~\ref{pangu_rmse_curve} shows the RMSE of MVAR\textsubscript{1hour} across all time steps. The sawtooth-like pattern in the RMSE curve arises from differences in accumulated errors at different time steps. Taking the RMSE curve of PM\textsubscript{2.5} as an example, there is a noticeable drop at the 24th hour compared to the previous time step because the 24-hour prediction is obtained directly from the 24-hour prediction model, involving only a single instance of error. In contrast, the prediction at the 23rd hour is derived from executing the 6-hour prediction model three times, the 3-hour prediction model once, and the 1-hour prediction model twice, resulting in a total of six instances of error.

\subsubsection{Pollutants Evolution Process Visualization}
\begin{figure*}[t] 
\centering
\includegraphics[width=\textwidth]{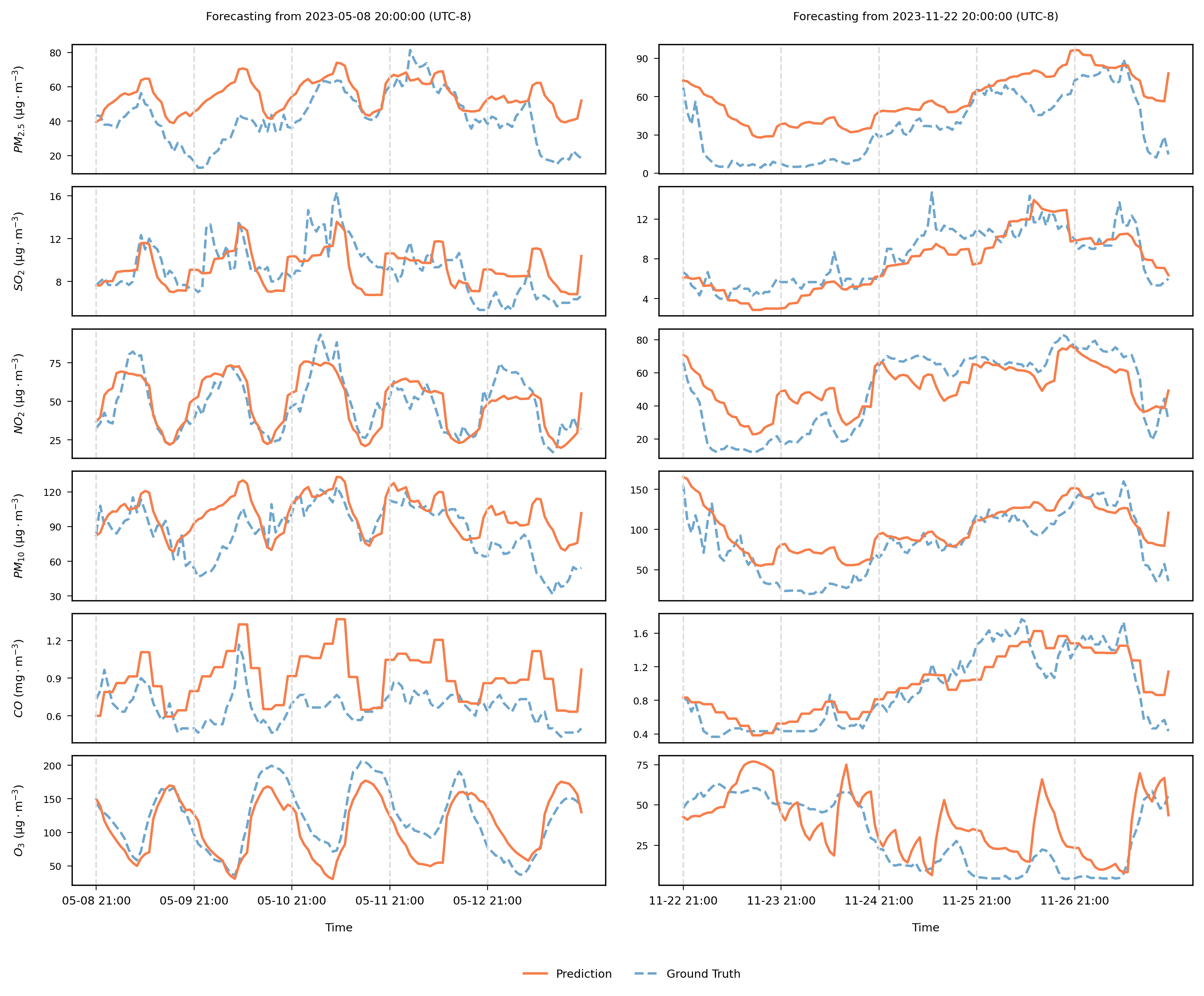}
\caption{Two pollutants evolution processes.} 
\label{pangu_pollutant_evolution_processes} 
\end{figure*}

To assess the feasibility of the 1-hourly modeling approach MVAR\textsubscript{1hour}, we analyze two pollution events, O\textsubscript{3} and PM\textsubscript{2.5}, across three cities, \ie, Beijing, Tianjin, and Langfang. The O\textsubscript{3} pollution event is initiated for forecasting at 20:00 (UTC-8) on May 8, 2023, and the PM\textsubscript{2.5} pollution event is initiated for forecasting at 20:00 (UTC-8) on November 22, 2023. Both events forecast for 120 hours at a 1-hour resolution. The average forecasted concentrations from the three cities are compared with the average observed concentrations. The results are shown in Figure~\ref{pangu_pollutant_evolution_processes}.

The first column presents the O\textsubscript{3} pollution event. From the observed curve, O\textsubscript{3} exhibits a clear diurnal cycle, with lower concentrations at night and a peak in the afternoon. The daily O\textsubscript{3} peak gradually increases starting from May 8, reaching its highest value by May 11. MVAR\textsubscript{1hour} effectively captures the diurnal variation of ozone, with strong consistency in predicting the timing of O\textsubscript{3} concentration peaks. NO\textsubscript{2}, which participates in atmospheric photochemical reactions with O\textsubscript{3}, is also well captured by MVAR\textsubscript{1hour} in terms of concentration changes. However, forecast accuracy decreases as the forecast time increases. During this pollution event, MVAR\textsubscript{1hour} successfully captures the overall trend of other pollutants and provides a good prediction of the peak concentration timing.

The second column presents a comparison of the PM\textsubscript{2.5} pollution event. Starting from the forecast initiation, PM\textsubscript{2.5} and PM\textsubscript{10} concentrations first decrease, gradually rise after 19:00 on November 23, 2023, peak in the early hours of November 27, and then rapidly decline. The trend in PM\textsubscript{*} concentrations aligns closely with SO\textsubscript{2}, CO, and NO\textsubscript{2}. MVAR\textsubscript{1hour} provides reliable trend forecasts for PM\textsubscript{*}, accurately predicting the peak values, which are particularly important in real-world applications. Notably, the forecast for PM\textsubscript{*} during the 110-120 hour window effectively captures the rapid decline in concentrations. Throughout this pollution event, MVAR\textsubscript{1hour} also performs well in forecasting SO\textsubscript{2}, CO, and NO\textsubscript{2}.

\end{document}